\documentclass[journal,comsoc]{IEEEtran}
\IEEEoverridecommandlockouts
\usepackage{cite}
\usepackage{amsmath,amssymb,amsfonts}
\usepackage{algorithmic}
\usepackage{graphicx}
\usepackage{textcomp}
\usepackage{xcolor}
\usepackage{float}
\usepackage{amsmath,amssymb}
\usepackage{bm}
\usepackage{bbm}
\usepackage{multirow} 
\usepackage{caption}
\usepackage[ruled,vlined]{algorithm2e}
\usepackage{booktabs}
\usepackage{array}
\usepackage{subfig}
\usepackage{rotating}

\newcommand{\vect}[1]{\mathbf{#1}}
\newcommand{\matr}[1]{\mathbf{#1}}

\newcommand{\set}[1]{\mathcal{#1}}

\def\BibTeX{{\rm B\kern-.05em{\sc i\kern-.025em b}\kern-.08em
		T\kern-.1667em\lower.7ex\hbox{E}\kern-.125emX}}
\usepackage[T1]{fontenc}

\begin{document}

\title{Weakly Supervised 3D Point Cloud Segmentation via Multi-Prototype Learning}

\author{Yongyi~Su$^\#$,
        Xun~Xu$^\#$,~\IEEEmembership{Senior Member,~IEEE,}
        and~Kui~Jia$^*$
\thanks{Yongyi~Su$^\#$ and Xun~Xu$^\#$ contributed equally to this work. Corresponding author: Kui~Jia$^*$. e-mail: kuijia@scut.edu.cn.}
\thanks{Xun~Xu is with Institute for Infocomm Research, A*STAR. Yonyi~Su and Kui~Jia are with South China University of Technology.}%
}

\maketitle
\begin{abstract}

Addressing the annotation challenge in 3D Point Cloud segmentation has inspired research into weakly supervised learning. Existing approaches mainly focus on exploiting manifold and pseudo-labeling to make use of large unlabeled data points. A fundamental challenge here lies in the large intra-class variations of local geometric structure, resulting in subclasses within a semantic class. In this work, we leverage this intuition and opt for maintaining an individual classifier for each subclass. Technically, we design a multi-prototype classifier, each prototype serves as the classifier weights for one subclass. To enable effective updating of multi-prototype classifier weights, we propose two constraints respectively for updating the prototypes w.r.t. all point features and for encouraging the learning of diverse prototypes. Experiments on weakly supervised 3D point cloud segmentation tasks validate the efficacy of proposed method in particular at low-label regime. Our hypothesis is also verified given the consistent discovery of semantic subclasses at no cost of additional annotations.

\end{abstract}


\IEEEpeerreviewmaketitle

\section{Introduction}
\begin{figure}
    \centering
    \includegraphics[width=1\linewidth]{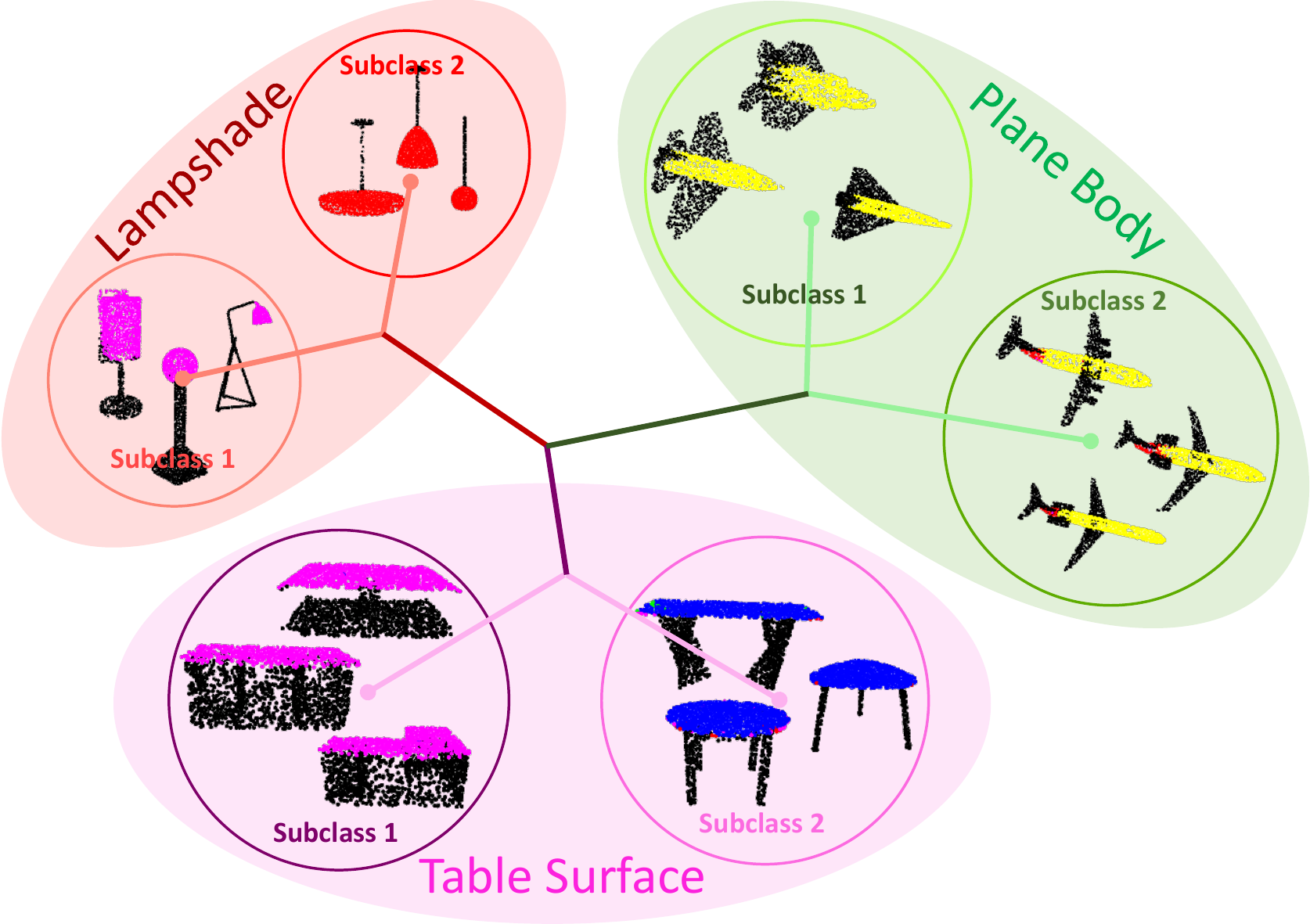}
    \caption{Illustration of the subclass concept. Colored points indicate the activated prototype/classifier, a clear subclass structure is observed from shape part categories, e.g. ``lampshade'' of pendants and lamps standing on the ground, ``plane body'' of fighters and passenger jets and ``table surface'' of square and round tables.}
    \label{fig:HeadPic}
    \vspace{-0.3cm}
\end{figure}

\IEEEPARstart{A}{nnotating} 3D point cloud for segmentation is expensive as it requires extensively annotating large number of points in 3D space and the 3D characteristics, e.g. occlusion, etc., render annotation particularly harder. A recent approach towards tackling the annotation challenge is through learning from partially labelled data, a.k.a. weakly supervised learning~\cite{xu2020weakly}. This initial attempt provided insight into the mechanism of weakly supervised learning for 3D semantic segmentation using the central limit theorem. They further proposed to strengthen the task by learning geometric manifold and consistency based semi-supervised learning. Based on these insights, follow-up works are carried out by introducing propagation methods to produce better pseudo-labels as supervision ~\cite{liu2021one,tao2020seggroup}. Despite the efforts on label propagation for more efficient use of limited labels, one inherent challenge in 3D point cloud segmentation, the large intra-class variation, remains unnoticed. For example, as illustrated in Fig.~\ref{fig:HeadPic}, there is a substantial diversity of visual appearance for ``plane body'', ``lampshade'' and ``table surface'' even if they respectively represents a single semantic category. This intra-class variation results in subclasses clearly identified within each semantic category. When a linear classifier with cross-entropy loss is applied, data points with the same label are forced to group together and center around a \textit{prototype} which is the weight of classifier~\cite{boudiaf2020unifying}. As a consequence, this requires a very complicated representation function to map data points with varying appearance to a single point in the feature space. However, when labeled data is extremely low, e.g. only 1 point per category, training a single linear classifier (prototype) for each category is prone to underfitting~\cite{allen2019infinite}.


Instead of having a single classifier/prototype for each semantic class, keeping multiple prototypes~\cite{allen2019infinite} has been adopted for few-shot learning to address multi-modal distribution within a semantic class. In this related problem, IMP~\cite{allen2019infinite} dynamically increases the number of prototypes following a Chinese restaurant process (CRP), when data points are far enough (above a threshold) to existing prototypes a new cluster center (prototype) is created. Despite the flexibility in determining the number of prototypes, CRP is essentially a sequential process and is only effective with a small support set in few-shot learning. Generalizing CRP to weakly supervised learning, where a large number of data (up to millions of points in a single mini-batch) determines the prototypes, is subject to prohibitive computation cost. In light of this challenge, we propose to introduce a multi-prototype classifier for weakly supervised segmentation, termed MulPro for simplicity. In specific, we first design a multi-prototype {memory bank} to store the prototypes for each semantic class and each prototype would represent one subclass. In contrast to the offline prototype updating with K-means clustering~\cite{rippel2015metric} or moving average update adopted by~\cite{liu2021one}, our design does not introduce any non-differentiable operations between prototypes and loss functions, thus enables end-to-end training. 

With the introduction of multi-prototype {memory bank}, a new challenge arises as how to effectively train the prototypes. One naive way would be direct backpropagating from cross-entropy loss on labeled data. However, this will only provide very sparse supervision signal when label data is few. To tackle this issue, we propose to use both labeled and unlabeled data. Given the assignment of points to a prototype we enforce the prototype to be close to the feature of all assigned points, resembling taking the average, thus we name also this constraint as subclass averaging constraint. Subclass averaging is differentiable w.r.t.  prototypes and can be used for gradient-base updating of prototypes. This whole process can interpreted as nesting K-means clustering within the classifier, the forward pass computes the assignment and cluster centers are updated by subclass averaging in the backward update. Both steps can be efficiently conducted in a single forward-backward iteration.

We further notice that the multi-prototype update through subclass averaging does not prevent degenerate prototypes, e.g. some or all prototypes in one class become identical. Such a solution will cause random activation of prototypes, which is harmful for gradient-based updating. Therefore, we further propose to force the prototypes to be diverse within a category. This is achieved by penalizing the similarity between prototypes and introducing a constraint balancing prototype diversity and separability.

Finally, we carry out experiments on weakly supervised segmentation for 3D point cloud datasets and demonstrate the effectiveness of multi-prototype classifier. We reveal that it is particularly useful when intra-class variation is large and labeled data is few. We are also surprised to see subclasses being discovered simultaneously with weakly supervised training without requiring extra fine-grained annotation.
The contributions of this work can be summarized as below.
\begin{itemize}
    \item Observing the large intra-class variation and clear subclass structures in 3D point cloud segmentation data, we propose a multi-prototype classifier, termed as MulPro, to reduce the difficulty in representation learning when only sparse labeled data is available.
    \item We propose a {subclass averaging} constraint to exploit both labeled and unlabeled data to supervise prototypes learning. This can be interpreted as nesting K-means clustering within the classifier. We further propose additional constraints to encourage diversity between prototypes to avoid degenerate solutions.
    \item We improve weakly supervised 3D point cloud segmentation tasks and simultaneously discover subclasses within each semantic categories without any additional supervision.
\end{itemize}

\section{Related Work}
\subsection{3D Point Cloud Segmentation}
Segmentation is a fundamental task in understanding 3D environment. The recent surge of deep learning methods for 3D point cloud is attributed to  PointNet\cite{Qi2017} which adapted a MLP to learn keypoints for point cloud understanding. The follow-up work, PointNet++~\cite{qi2017pointnet++},  proposed to embed PointNet in a small neighborhood to capture  local geometric information. Convolution based methods~\cite{li2018pointcnn, Wu_2019_CVPR, wang2018deep, 9010002} and Graph convolution based methods~\cite{wang2019dynamic, Landrieu_2018_CVPR, 9010334, 9051667} are subsequently proposed to further expand the receptive field and learn better local geometry. More recently, transformers~\cite{zhao2021point,guo2021pct} are adapted to 3D point cloud, achieving unprecedented performance. A more detailed review of point cloud understanding can be found in \cite{guo2020deep}. Nevertheless, the success of 3D point cloud segmentation is mainly attributed to training on large amount of labeled data. While labeling on 3D data for segmentation is particularly expensive and it has inspired works addressing weakly-supervised learning for 3D point cloud segmentation.

\subsection{3D Point Cloud Weakly Supervised Learning.}
Annotating 3D data for segmentation is expensive due to the high degree of freedom and articulated boundaries. Weakly supervised learning addresses this issue by exploiting sparsely labeled data~\cite{meng2021towards,wei2020multi, xu2020weakly,liu2021one,tao2020seggroup}. \cite{wei2020multi} proposed a inexact annotation scheme by providing multiple binary labels to one region, class activation map (CAM)~\cite{zhou2016learning} is employed to exploit these labels to infer point-wise predictions. In another line of research, \cite{xu2020weakly} assumes only a fraction of points are uniformly selected and labeled. They proved that under i.i.d. assumption, the weakly supervised learning gradient will approximate the fully supervised one. Inspired by the discovery made in \cite{xu2020weakly}, a pseudo-labeling approach \cite{liu2021one} further improved label propagation to better exploit the unlabeled points. In \cite{liu2021one}, a per-region annotation assumption is adopted which exploits the unique information provided by ScanNet dataset.
The strong assumption makes \cite{liu2021one} restrictive to particular datasets where perfect super-point region is provided. More recently, motivated by saving annotation cost, an active learning approach towards point cloud data \cite{shi2021label} is investigated to select most informative points or super-points to annotate. Alternative to mining the geometric properties, PointContrast~\cite{xie2020pointcontrast} proposed an unsupervised contrastive pretraining approach to adjust model parameters on large unlabeled data. Finetuning on small label data demonstrates promising results on label-efficient learning.
As opposed to propagating labels and pretraining, in this work, we are motivated by the large intra-class variation and concluded that having multiple prototypes/classifiers for each semantic class could alleviate the difficulty in representation learning. 

\subsection{Multi-Prototype Learning.}
Originated in few-shot learning, multi-prototype learning aims to address the challenging of fitting prototypical network~\cite{snell2017prototypical} for multi-modal data distribution~\cite{zhao2021few, allen2019infinite, Deuschel_2021_ICCV} by learning prototypes for recognizing classes with few training examples. The first attempt, IMP~\cite{allen2019infinite}, proposed to adaptively expand prototype pool following a Chinese restaurant process which sequentially processes data points. \cite{Deuschel_2021_ICCV} proposes a k-means extension of Prototypical Networks. Despite the success in few-shot learning, it is impractical to trivially apply the sequential IMP to weakly supervised learning due to computation cost while other offline methods prevents end-to-end training of prototypes. 
In this work, we develop a multi-prototype memory bank to capture the subclass structure and the proposed constraints allow effective multi-prototype training.
In the context of 3D point cloud deep learning, exploiting multi-prototype was briefly mentioned in \cite{zhao2021few} which generates multiple prototypes by farthest point sampling on the embedding space for support set. Compared with \cite{zhao2021few}, we provide the first in-depth analysis into multi-prototype in weakly supervised learning. The key novelty is how to effectively learn non-trivial multiple prototypes and demonstrating its existence. Without the introduced class averaging constraint and diversity constraints, as shown in the ablation, multi-prototype would not be effective. In contrast, multi-prototype is implemented as clustering support points and no concrete evidence of the existence of multiple prototypes is provided in \cite{zhao2021few}.

\section{Methodology}
We introduce a multi-prototype classifier, MulPro, to exploit sparse annotations. In this section, we first formally define the weakly supervised segmentation task. Then, we describe how the multi-prototype classifier works in the weakly supervised model. Finally, in view of the difficulties in learning multiple prototypes we propose two constraints to further constrain the prototype representation learning.

\subsection{Architecture Overview}


To formally define the weakly-supervised 3D point segmentation task, we follow the settings proposed in ~\cite{xu2020weakly}. In specific, a training dataset $\set{D}_{tr}=\{\matr{X}_i,\matr{Y}_i,\matr{M}_i\}_{i=1\cdots N_{tr}}$ is provided, where $\matr{X}_i\in \mathbb{R}^{D_i\times N}$ is the input $N$ point features, e.g. 3D coordinates with RGB color if available, $\matr{Y}_i\in \{0,1\}^{K\times N}$ is the one-hot per-point segmentation label ($K$ categories) and $\matr{M}_i\in\{0,1\}^{N}$ is a binary mask indicating whether ground-truth label is available. An encoder network $\matr{Z}=f(\matr{X};\Theta)$ maps input points into a feature space $\matr{Z}\in\mathbb{R}^{D_o\times N}$. A classifier $g(\matr{Z};\matr{\Omega})\in\mathbb{R}^{K\times N}$ maps encoded features into logits in segmentation category space. The existing approaches~\cite{xu2020weakly, liu2021one} often define cross-entropy loss on classifier outputs and additional regularization may derive from manifold~\cite{xu2020weakly}, pseudo-labeling~\cite{liu2021one}, etc. 
In contrast to the above approaches, we propose to improve the classifier layer by introducing multiple prototypes to exploit the underlying subclass structures. In specific, the linear classifier $g(\matr{Z};\matr{\Omega})$ can be expressed as below if bias is removed,
\begin{equation}
  g(\matr{Z};\Omega)=\vect{\Omega}^\top \matr{Z},\quad s.t.\quad\vect{\Omega}\in\mathbb{R}^{D_o\times K}
\end{equation}

\begin{figure*}[!htb]
\centering
\includegraphics[width=0.88\linewidth]{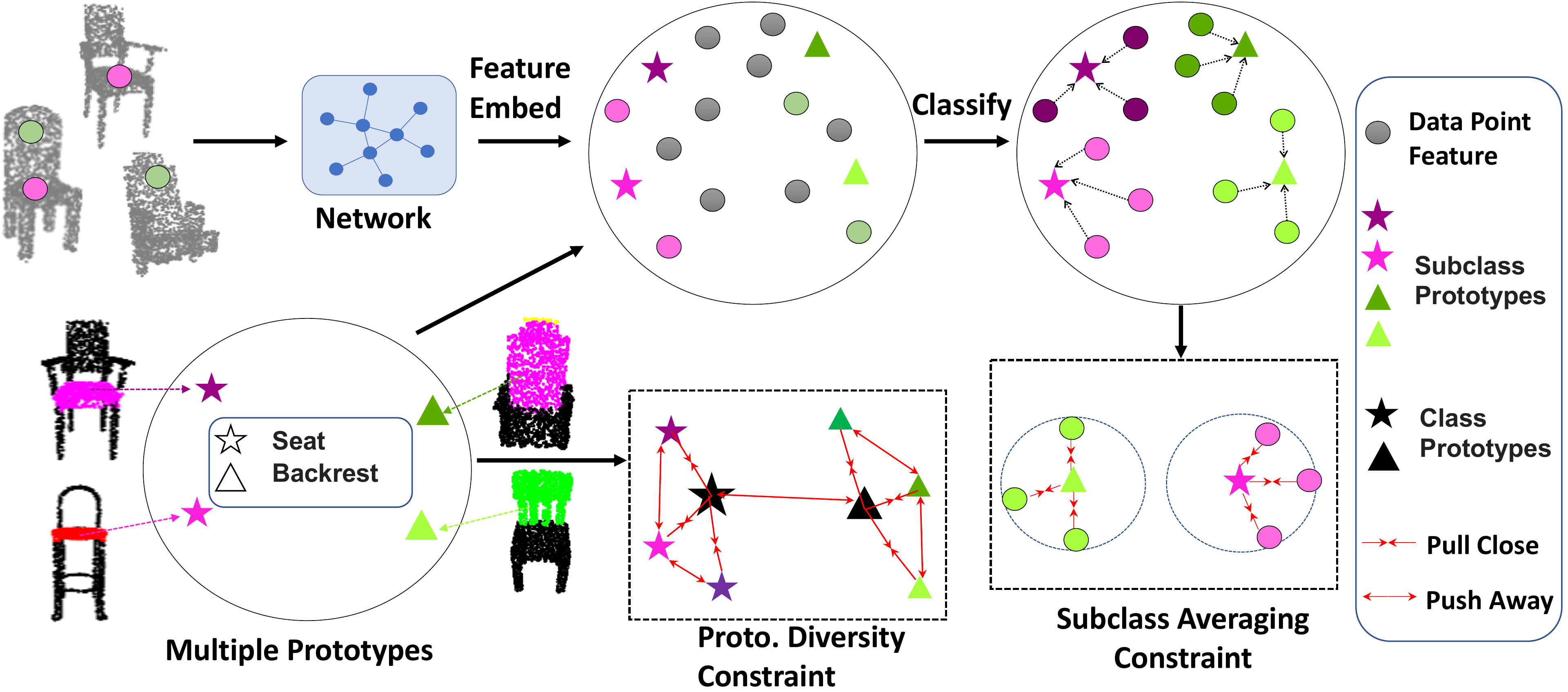}
\caption{An overview of MulPro for weakly supervised learning. With multiple prototypes per class (two stars and two triangles represent different types of seat and backrest respectively) data points are classified by the closest prototype. Subclass averaging and prototype diversity constraints are employed to learn multi-prototypes effectively.
}\label{fig:overview}
\end{figure*}

Training the linear classifier with cross-entropy loss can be seen as discovering $K$ prototypes, each represented as $\vect{\omega}_k\in\mathbb{R}^{D_o}$, in the encoded feature space such that points belonging to the same category group together and center around the prototype, while pushing different categories away~\cite{boudiaf2020unifying}. With such a design, it is assumed that a single prototype $\vect{\omega}_k$ is discovered for each semantic category. However, observing subclasses within each semantic category, the single prototype assumption could be too strong and potentially increases the risk of underfitting when small labeled data is available. To tackle this issue, we introduce a multi-prototype memory bank which maintains multiple prototypes for each class and each prototype accounts for one subclass. An overview of MulPro is illustrated in Fig.~\ref{fig:overview}.

    
\subsection{Multi-Prototype Classifier}
We define a multi-prototype memory bank as $\matr{\Omega}\in\mathbb{R}^{D_o\times M\times K}$. In the forward pass, given encoded feature $\matr{Z}$, we first take the inner product with all prototypes and this results in an attention map $\matr{A}\in\mathbb{R}^{N\times M\times K}$ as below.
\begin{equation}\label{eq:InnerProd}
    a_{nkm} = \sum_d{z_{dn} \cdot \omega_{dmk}}
\end{equation}

We notice that when $M=1$ the multi-prototype classifier simply degenerates to standard linear classifier. With $M>1$, we apply a maxpooling operation on the attention map $\matr{A}$ and this yields the classification logits $l_{kn}=\max_m a_{nkm}$. The logits are eventually used for calculating the cross-entropy loss as,
\begin{equation}\label{eq:CEloss}
    \mathcal{L}_{CE}= -\frac{1}{N}\sum_n\sum_ky_{kn}\log \frac{\exp(l_{kn})}{\sum_k\exp(l_{kn})}
\end{equation}

\noindent\textbf{Discussion}.
We provide a few insights for the multi-prototype design here. First, in the forward pass, multi-prototype classifier acts like a $K\times M$ way classifier, while the linear classifier is a $K$-way classifier. Each data point feature is evaluated against all prototypes through Eq.~(\ref{eq:InnerProd}). Instead of directly classifying into one of the $K\times M$ classes, to respect the label ground-truth being $K$-way, the multi-prototype classifier first reduces the $K\times M$-way prediction into $K$-way prediction through maxpooling over prototypes (the $M$ dimension). As a result, $K\times M$ prototypes can be learned with only $K$-way labels provided. 
Because of the subclass structures, through this design we can identify the subclasses (up to $M$) as training goes on. Each prototype will naturally represent the subclass center.





\begin{figure}
    \centering
    \includegraphics[width=1.01\linewidth]{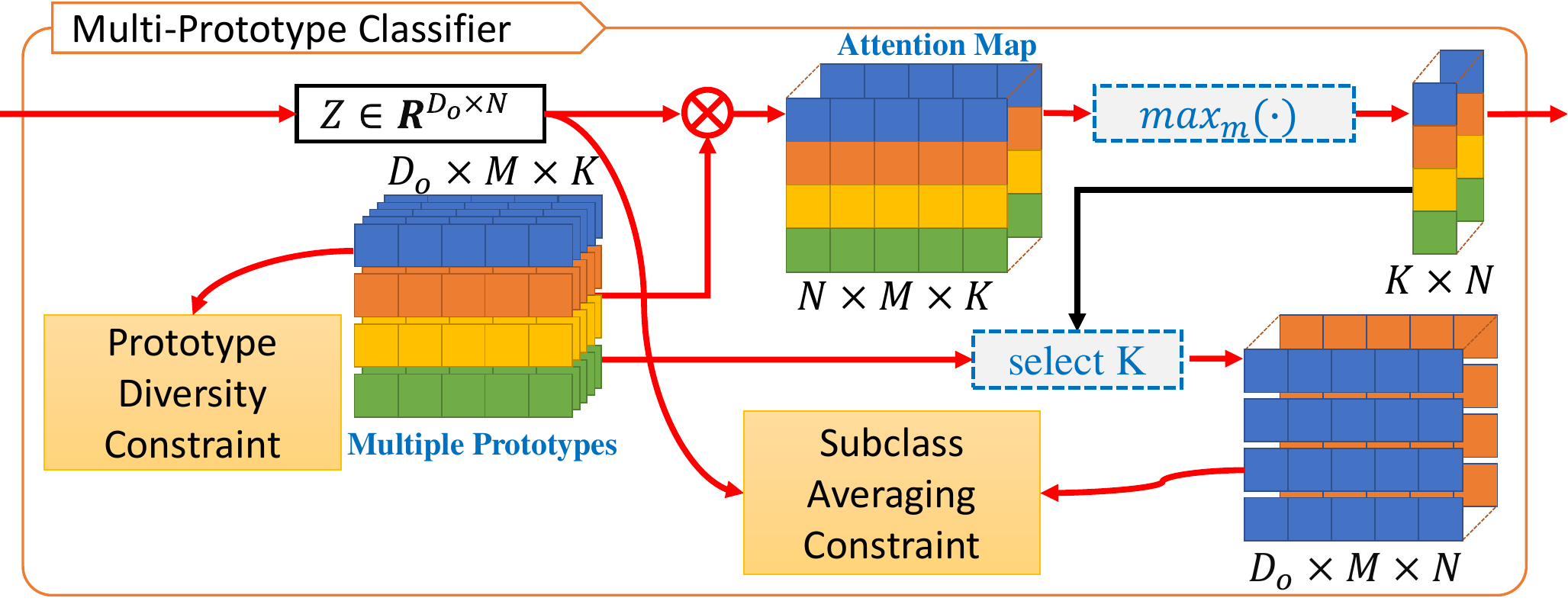}
    \caption{Multi-prototype classifier schematic. Red lines indicate back propagation flow.}
    \label{fig:MultiPrototypeClassifier}
\end{figure}

\subsection{Multi-Prototype Updating}

We now elaborate how the multi-prototypes are updated. We first denote the activated prototype $\vect{\omega}_{\hat{m}\hat{k}}$ as the following, 
\begin{equation}\label{eq:activeproto}
    \vect{\omega}_{\hat{m}_n\hat{k}_n}\in\mathbb{R}^{D_o}\quad s.t.\quad \hat{k}_n,\hat{m}_n=\arg\max_{k}\max_{m} a_{nkm}
\end{equation}

Given an activated prototype $\vect{\omega}_{\hat{m}\hat{k}}$, one could easily verify that the classification logit can be written as,
\begin{equation}
    l_{kn}=\vect{\omega}_{\hat{m}_n\hat{k}_n}^\top \vect{z}_n
\end{equation}

Therefore, each activated prototype can be directly updated by backpropagating from cross-entropy loss defined in Eq.~(\ref{eq:CEloss}). Nevertheless, sparse labelled data provides weak supervision over prototypes and additional regularization is necessary for learning high quality prototypes.
To this end, we further propose two constraints for training prototypes, namely {subclass averaging constraint} and {prototype diversity constraint}.

\subsubsection{Subclass Averaging Constraint}
The existing weakly supervised approaches update prototypes via  exponential moving averaging over labeled data points~\cite{liu2021one}. As a result, it prohibits learning from unlabeled data points. In this section, we introduce a differentiable loss on all available data to provide supervision signal in addition to the cross-entropy loss. 



{Specifically, we first identify the per-class activated prototype for each point as $\Omega_{\hat{k}_n} = [\vect{\omega_{1\hat{k}_n}}; \cdots ; \vect{\omega_{m\hat{k}_n}}]$. These prototypes are further stacked as,}
\begin{equation}
    \widetilde{\Omega}=[\Omega_{\hat{k}_1}; \cdots ;\Omega_{\hat{k}_N}] \in \mathbb{R}^{D_o\times M \times N}
\end{equation}

{The subclass averaging constraint $\mathcal{L}_{avg}$ is then implemented as in Alg.~\ref{alg:Recon}. This algorithm accomplishes two tasks: activating the prototypes and updating the activated prototypes. We use a threshold $\gamma$ to determine whether the data point feature belongs to one prototype. If one data point feature is similar to a prototype, above the threshold $\gamma$, $\mathcal{L}_{avg}$ pulls them closer. Otherwise, we use it to update (activate) the farthest prototype. Therefore, the cosine distance $1-\gamma$ represents the radius of a hypersphere covering the data points that are used to update the prototype in the center. More empirical analysis into the subclass averaging constraint design is presented in Sect.~\ref{subsection:ReconLossDesign}.} 

\begin{algorithm}[hb]
\SetKwInOut{Input}{input}
\SetKwInOut{Output}{output}
\Input{\small{Point-wise encoded features $\matr{Z}\in \mathbb{R}^{N\times D_o}$
Prototypes within the category of $\omega_{\hat{m}_n\hat{k}_n}$ as $\widetilde{\Omega} \in \mathbb{R}^{D_o \times M \times N}$
}}

\Output{\small{{Subclass averaging} loss $\mathcal{L}_{avg}$}}

calculate cosine similarity between point-wise features and the prototypes as $\matr{S}\in \mathbb{R}^{N\times M}$, $s_{nm} = \frac{\widetilde{\omega}_{mn}^\top \vect{z}_n}{||\vect{z}_n|| \cdot ||\widetilde{\omega}_{mn}||}$

initialize a $\matr{0}$-matrix $\matr{W}\in \{0\}^{N\times M}$

\For{$n \leftarrow 1$ \KwTo $N$}
{
    \If{$\max_m{s_{nm}} > \gamma$}
    {
        $w_{n} = softmax(\vect{s}_n / \tau)$
    }
    \Else
    {
        $w_{n} = softmax(-\vect{s}_n / \tau)$
    }
}

$\mathcal{L}_{avg} = - \sum_n\sum_m{w_{nm} \cdot s_{nm}}$

\Return $\mathcal{L}_{avg}$

\caption{\small{{Subclass averaging constraint} algorithm}\label{alg:Recon}}
\end{algorithm}

\noindent\textbf{Discussion}.
The proposed {subclass averaging} constraint can be interpreted as using data point features to update prototypes. Compared with moving average update adopted in \cite{liu2021one}, our design is superior in two ways. First, it is differentiable and can be combined with other learning objectives for an end-to-end training, while moving average update prevents combination with other losses to update the prototypes. Second, it allows using all data points, both labeled and unlabeled, to update prototypes. This enables discovering subclasses from all available data. Finally, since cosine similarity is agnostic to scale, minimizing $\mathcal{L}_{avg}$ will not result in an explosion of scale.

\subsubsection{Prototype Diversity Constraint}\label{sec:PDC}
Empirical results from the experiment suggest that directly updating the multi-prototype with cross-entropy loss and {subclass averaging} constraint does not necessarily guarantee all subclasses being discovered. In another words, there is a risk of all prototypes within a semantic class collapsing into an identical one. To avoid the collapsing issue, we further propose  prototype diversity constraints to encourage  diverse prototypes being discovered. 

\noindent\textbf{Prototype Diversity within a Semantic Class}.
First of all, to encourage more diverse prototypes within a semantic class we penalize the accumulative similarity between prototypes as,
\begin{equation}\label{eq:protodiv}
    \mathcal{L}_{pd}=\sum_{k=1}^K\sum_{m_i=1}^M\sum_{m_j=1}^M sim(\omega_{m_i,k},\omega_{m_j,k})
\end{equation}

For the selection of similarity metric $sim$, we take the following considerations into account. First, the similarity should avoid any trivial solution. Therefore any unconstrained similarity metrics should be excluded, e.g. inner product could result in an vanish of scale. Moreover, the diversity should not be overly emphasized, otherwise all prototypes could become equally distanced and they no longer characterize the subclasses within a single class. As a result, we propose to adopt a piece-wise similarity function, specifically the follow thresholded cosine similarity is adopted,
\begin{equation}
    sim=\max(0,\frac{\omega_{m_i,k}^\top\omega_{m_j,k}}{||\omega_{m_i,k}||\cdot||\omega_{m_j,k}||}-\sigma)
\end{equation}

This indicates prototype diversity within a semantic class is encouraged only when the similarity is above a threshold $\sigma$ and the normalized cosine similarity avoids scale vanishing.

\noindent\textbf{Balancing Prototype Diversity and Class Separability}.
As discussed, overly emphasizing prototype diversity could result in equally distanced prototypes and harm the separability in semantic classes. To address this issue, we introduce a metric learning loss~\cite{xu2021learning} to apply further constraints to the distribution of prototypes. We first denote the per-class mean prototype $\bar{\omega}_k$ as the average of all normalized prototypes within each semantic class, i.e. $\bar{\omega}_k=\frac{1}{M}\sum_m \omega_{mk}/||\omega_{mk}||_2$, and the scatterness of prototypes as the variance, $s_k=\frac{1}{M}\sum_m ||\omega_{mk}/||\omega_{mk}||_2-\bar{\omega}_k/||\bar{\omega}_k||_2||^2_2$. We define the constraint as minimizing the negative logarithm of the ratio between minimal inter-class mean prototype distance and maximal intra-class scatterness as in Eq.~(\ref{eq:MIMI}).

\begin{equation}\label{eq:MIMI}
\begin{split}
    \mathcal{L}_{bds} =& -\log\frac{\min\limits_{k_i,k_j}||\frac{\bar{\omega}_{k_i}}{||\bar{\omega}_{k_i}||_2}-\frac{\bar{\omega}_{k_j}}{||\bar{\omega}_{k_j}||_2}||^2_2}{\max_{k} s_k}\\
    =&-\log\min_{k_i,k_j}||\frac{\bar{\omega}_{k_i}}{||\bar{\omega}_{k_i}||_2}-\frac{\bar{\omega}_{k_j}}{||\bar{\omega}_{k_j}||_2}||^2_2 + \log\max_{k} s_k
\end{split}
\end{equation}

\subsection{Training Strategy}
Finally, we combine cross-entropy loss, {subclass averaging} constraint and prototype diversity constraint to supervise the update of multi-prototype memory bank.

\begin{equation}
    \mathcal{L}=\lambda_{CE}\mathcal{L}_{CE} +\lambda_{avg}\mathcal{L}_{avg} + \lambda_{pd}\mathcal{L}_{pd} + \lambda_{bds}\mathcal{L}_{bds}
\end{equation}

We further notice that MulPro is compatible with the additional constraints and post-processing techniques introduced in \cite{xu2020weakly}.

\section{Experiment}
prototypes on ShapeNet datasetIn this section, we first introduce the benchmark datasets (Sect. \ref{subsection:datasets}). Next, we present details on our weakly supervised semantic segmentation experiments and compare with state-of-the-art methods (Sect. \ref{subsection:WSSS}). 
{We further provide analysis about multi-prototype} (Sect. \ref{subsection:multiPrototypeClassifierAnalysis}).
Finally, the ablation study demonstrates the superiority of the multi-prototype classifier and the importance of the several losses that constrain the multi-prototype update (Sect. \ref{subsection:ablationStudy}).

\subsection{Datasets} \label{subsection:datasets}
We conduct experiments on two 3D point cloud segmentation datasets.

\noindent\textbf{ShapeNet}\cite{yi2016scalable} is a richly-annotated, large-scale CAD model dataset including 16,881 shapes, divided into 16 categories, each annotated with 50 parts. We evaluate part segmentation task on this dataset following the weakly supervised setting proposed in \cite{xu2020weakly}. For each training sample, a subset of points, 10 percent of all points or one point per part, are randomly selected to be labelled. For testing, the comparison is performed using the default protocol.

\noindent\textbf{S3DIS}\cite{Armeni_2016_CVPR} is an indoor real scene dataset, which is widely used as a benchmark dataset for 3D segmentation evaluation. It is composed of 6 areas each including several rooms, e.g. office areas, educational and exhibition spaces. For scene segmentation, it has 13 semantic categories of indoor scene objects. Each point is represented with xyz coordinate and RGB value. For weakly supervised supervised setting~\cite{xu2020weakly}, a subset of points are uniformly labelled within each room. We choose Area 5 to be the test split.

\noindent\textbf{PartNet}\cite{mo2019partnet} is a CAD model dataset with 24 shape categories and a total of 26,671 unique objects. For part segmentation task, we choose the most coarse level annotation. The experiment setting is kept the same with \cite{xu2020weakly}.

\subsection{Weakly Supervised Semantic Segmentation} \label{subsection:WSSS}

\noindent\textbf{Encoder Network.} We use the feature extraction encoder of DGCNN\cite{wang2019dynamic} with default parameters combined with our Multi-Prototype classifier as our network for fair comparison with previous work~\cite{xu2020weakly}. 
{Here we set $\lambda_{CE}=\lambda_{avg}=\lambda_{pd}=\lambda_{bds}=1$, the hyper-parameter threshold $\sigma$ in prototype diversity constraint to 0.2 or 0.8 for different experiments, the hyper-parameter $\gamma$ in subclass averaging constraint algorithm to $\cos{\frac{\arccos{\sigma}}{2}}$ and the $\tau$ to $0.1$. $M$ is set as 5 for ShapeNet, and 10 for S3DIS to obtain the best results.}

\begin{table*}[htbp]
  \centering
 \caption{{mIoU ($\%$) evaluation on ShapeNet dataset. The fully supervision (Ful. Sup.) methods are trained on $100\%$ labelled points. Two levels of weak supervisions (1pt and $10\%$) are compared. The results presented here are with post-processing introduced in~\cite{xu2020weakly}.}}
   \setlength\tabcolsep{2pt} 
  \resizebox{0.99\linewidth}{!}{
      \begin{tabular}{cclcccccccccccccccccc}
    \toprule
    \multicolumn{2}{c}{Setting} & \multicolumn{1}{c}{Model} & SampAvg & CatAvg & Air.  & Bag   & Cap   & Car   & Chair & Ear.  & Guitar & Knife & Lamp  & Lap.  & Motor. & Mug   & Pistol & Rocket & Skate. & Table \\
    \midrule
    \multicolumn{2}{c}{\textbf{\footnotesize{Ful.Sup.}}} 
    & DGCNN\cite{wang2019dynamic} & 85.1 & 82.3 & 84.2 & 83.7 & 84.4  & 77.1  & 90.9 & 78.5 & 91.5 & 87.3 & 82.9  & 96.0 & 67.8  & 93.3  & 82.6 & 59.7 & 75.5  & 82.0 \\
    \midrule
    \multirow{10}[3]{*}{\begin{sideways}\textbf{Weak. Sup.}\end{sideways}} & \multirow{5}[1]{*}{\begin{sideways}1pt\end{sideways}} 
    & $\Pi$ Model\cite{Laine_ICLR17} & 73.2 & 72.7 & 71.1 & 77.0 & 76.1 & 59.7 & 85.3 & \textbf{68.0} & 88.9 & 84.3 & 76.5 & \textbf{94.9} & 44.6 & 88.7 & 74.2 & 45.1 & 67.4 & 60.9 \\
    &   & MT\cite{tarvainen2017mean} & 72.2 & 68.6 & 71.6 & 60.6 & 79.3 & 57.1 & 86.6 & 48.4 & 87.9 & 80.0 & 73.7 & 94.0 & 43.3 & 79.8 & 74.0 & 45.9 & 56.9 & 59.8 \\
    &   & WeakSup~\cite{xu2020weakly} & 75.5  & 74.4 & 75.6 & 74.4 & 79.2 & 66.3 & 87.3 & 63.3 & 89.4 & 84.4 & 78.7 & 94.5 & 49.7 & 90.3 & 76.7 & 47.8 & 71.0 & 62.6 \\
    &   & OTOC*~\cite{liu2021one} & 75.5 & 74.3 & 75.0 & 74.0 & \textbf{85.6} & 63.2 & 88.2 & 58.0 & 89.0 & 85.1 & 80.6 & 94.6 & 48.4 & 91.6 & 72.2 & 49.8 & 70.7 & 62.2 \\
    &   & MulPro (Ours) & \textbf{79.4} & \textbf{77.8} & \textbf{77.6} & \textbf{80.2} & 84.2 & \textbf{69.6} & \textbf{88.9} & 65.3 & \textbf{89.7} & \textbf{85.9} & \textbf{83.6} & 94.4 & \textbf{60.0} & \textbf{94.0} & \textbf{79.2} & \textbf{51.5} & \textbf{71.4} & \textbf{69.9} \\
          
\cmidrule{2-21}          & \multirow{5}[2]{*}{\begin{sideways}10\%\end{sideways}}
    & $\Pi$ Model\cite{Laine_ICLR17} & 83.8 & 79.2 & 80.0 & 82.3 & 78.7 & 74.9 & 89.8 & 76.8 & 90.6 & 87.4 & 83.1 & 95.8 & 50.7 & 87.8 & 77.9 & 55.2 & 74.3 & 82.7 \\
    &   & MT\cite{tarvainen2017mean} & 81.7 & 76.8 & 78.0 & 76.3 & 78.1 & 64.4 & 87.6 & 67.2 & 88.7 & 85.5 & 79.0 & 94.3 & 63.3 & 90.8 & 78.2 & 50.7 & 67.5 & 78.5 \\
    &   & WeakSup~\cite{xu2020weakly} & 85.0 & 81.7 & 83.1 & 82.6 & 80.8 & 77.7 & 90.4 & 77.3 & 90.9 & 87.6 & 82.9 & \textbf{95.8} & 64.7 & 93.9 & 79.8 & \textbf{61.9} & \textbf{74.9} & \textbf{82.9} \\
    &   & OTOC*~\cite{liu2021one} & 85.0 & \textbf{82.2} & \textbf{83.5} & \textbf{84.1} & 85.0 & 77.2 & \textbf{90.7} & 76.7 & 91.2 & \textbf{88.0} & 84.9 & 95.8 & 68.8 & 93.2 & \textbf{81.7} & 57.4 & 74.9 & 81.5\\
    &   & MulPro (Ours) & \textbf{85.3} & 82.0 & 83.3 & 80.0 & \textbf{85.2} & \textbf{77.7} & 90.5 & \textbf{77.5} & \textbf{91.2} & 87.9 & \textbf{85.1} & 95.6 & \textbf{69.4} & \textbf{94.4} & 81.2 & 56.4 & 73.8 & 82.7\\
    \bottomrule
    \end{tabular}%
    }
  \label{tab:ShapeNet}%
\end{table*}%

\begin{table*}[htbp]
  \centering
  \caption{{mIoU ($\%$) evaluations on S3DIS (Area 5) dataset. We compared against fully supervised (Ful.Sup.) and alternative weakly supervised (Weak. Sup.) approaches.}}
  \setlength\tabcolsep{2pt} 
	  \resizebox{0.8\linewidth}{!}{
    \begin{tabular}{cclcccccccccccccc}
    \toprule
    \multicolumn{2}{c}{Setting} & Model & \multicolumn{1}{l}{CatAvg} & \multicolumn{1}{l}{ceil.} & \multicolumn{1}{l}{floor} & \multicolumn{1}{l}{wall} & \multicolumn{1}{l}{beam} & \multicolumn{1}{l}{col.} & \multicolumn{1}{l}{win.} & \multicolumn{1}{l}{door} & \multicolumn{1}{l}{chair} & \multicolumn{1}{l}{table} & \multicolumn{1}{l}{book.} & \multicolumn{1}{l}{sofa} & \multicolumn{1}{l}{board} & \multicolumn{1}{l}{clutter} \\
    \midrule
    
    \multicolumn{2}{c}{\textbf{\footnotesize{Ful.Sup.}}}
        & DGCNN\cite{wang2019dynamic} & 47.0  & 92.4 & 97.6 & 74.5 & 0.5 & 13.3 & 48.0 & 23.7 & 65.4 & 67.0 & 10.7 & 44.0 & 34.2 & 40.0 \\
    \midrule
    \multirow{10}[3]{*}{\begin{sideways}\textbf{Weak. Sup.}\end{sideways}} & \multirow{5}[1]{*}{\begin{sideways}1pt\end{sideways}} 
        & $\Pi$ Model\cite{Laine_ICLR17} & 44.3 & 89.1 & 97.0 & 71.5 & 0.0 & 3.6 & 43.2 & 27.4 & 62.1 & 63.1 & 14.7 & \textbf{43.7} & 24.0 & 36.7 \\
    &   & MT\cite{tarvainen2017mean} & 44.4 & 88.9 & 96.8 & 70.1 & 0.1 & 3.0 & 44.3 & 28.8 & 63.6 & 63.7 & 15.5 & \textbf{43.7} & 23.0 & 35.8 \\
    &   & WeakSup~\cite{xu2020weakly} & 44.5 & 90.1 & \textbf{97.1} & \textbf{71.9} & 0.0 & 1.9 & \textbf{47.2} & 29.3 & 62.9 & 64.0 & 15.9 & 42.2 & 18.9 & 37.5 \\
    &   & OTOC*\cite{liu2021one} & 45.6 & 89.0 & 96.6 & 69.0 & \textbf{0.2} & \textbf{7.6} & 43.6 & 34.4 & 59.4 & 59.7 & 16.1 & 43.2 & \textbf{36.9} & 37.1 \\
    &   & MulPro~(Ours) & \textbf{47.5} & \textbf{90.1} & 96.3 & 71.8 & 0.0 & 6.7 & 46.7 & \textbf{39.2} & \textbf{67.2} & \textbf{67.4} & \textbf{21.8} & 39.2 & 33.0 & \textbf{38.0} \\
    
\cmidrule{2-17}          & \multirow{5}[2]{*}{\begin{sideways}10\%\end{sideways}} 
        & $\Pi$ Model\cite{Laine_ICLR17} & 46.3  & 91.8  & 97.1 & 73.8  & 0.0 & 5.1 & 42.0 & 19.6 & 66.7 & 67.2 & 19.1 & 47.9 & 30.6 & 41.3 \\
    &   & MT\cite{tarvainen2017mean} & 47.9 & \textbf{92.2} & 96.8 & 74.1 & 0.0 & 10.4 & 46.2 & 17.7 & 67.0 & 70.7 & \textbf{24.4} & 50.2 & 30.7 & 42.2 \\
    &   & WeakSup~\cite{xu2020weakly} & 48.0 & 90.9 & 97.3 & 74.8 & 0.0  & 8.4  & \textbf{49.3} & 27.3 & \textbf{69.0} & \textbf{71.7} & 16.5 & \textbf{53.2} & 23.3 & 42.8 \\
    &   & OTOC*~\cite{liu2021one} & 48.2 & 91.2 & \textbf{97.7} & \textbf{78.0} & 0.0 & 6.3 & 46.3 & 31.6 & 65.7 & 64.4 & 8.2 & 52.5 & \textbf{41.6} & \textbf{43.1} \\
    &   & MulPro~(Ours) & \textbf{49.0} & 89.7 & 96.9 & 75.5 & 0.0 & \textbf{14.0} & 45.7 & \textbf{40.7} & 68.5 & 66.8 & 13.9 & 49.4 & 34.4 & 41.2 \\
    \bottomrule
    \end{tabular}
    }
  \label{tab:S3DIS_PartNet}
\end{table*}

\begin{table*}[]
    \centering
    \caption{Detailed results on PartNet dataset at 1pt annotation cost. We report mIoU~($\%$) of segmentation for each object category.}
    \setlength\tabcolsep{1pt} 
    \resizebox{1\linewidth}{!}{
        \begin{tabular}{c|cccccccccccccccccccccccccc}
        \toprule
        \multicolumn{1}{c|}{Setting} & Model & CatAvg & Bag & Bed & Bott. & Bowl & Chair & Clock & Dish. & Disp. & Door & Ear. & Fauc. & Hat & Key. & Knife & Lamp & Lap. & Micro. & Mug & Frid. & Scis. & Stora. & Table & Trash. & Vase \\
        \midrule
        \multicolumn{1}{c|}{\textbf{Ful.Sup.}} & DGCNN & 65.6 & 53.3 & 58.6 & 48.9 & 66.9 & 69.1 & 35.8 & 75.2 & 91.2 & 68.5 & 59.3 & 62.6 & 63.7 & 69.5 & 71.8 & 38.5 & 95.7 & 57.6 & 83.3 & 53.7 & 89.7 & 62.6 & 65.3 & 67.8 & 66.8 \\
        \midrule
        \multirow{3}[3]{*}{\begin{sideways}\textbf{Weak Sup.}\end{sideways}}  & Baseline & 50.2 & 24.4 & 30.1 & 20.5 & 38.0 & 65.9 & 35.3 & 64.9 & {84.3} & {52.6} & 36.7 & 47.1 & 47.9 & 52.2 & 55.2 & 34.1 & 92.4 & 49.3 & 59.5 & 49.6 & 80.1 & 44.6 & 49.8 & 40.4 & 49.5 \\
         & WeakSup~\cite{xu2020weakly} & 54.6 & 28.4 & 30.8 & 26.0 & 54.3 & \textbf{66.4} & 37.7 & 66.3 & 81.0 & 51.7 & 44.4 & 51.2 & 55.2 & 56.2 & 63.1 & 37.6 & \textbf{93.5} & 49.7 & 73.5 & 50.6 & \textbf{83.6} & 46.8 & 61.1 & 44.1 & 56.8 \\
         & MulPro (Ours) & \textbf{60.9} & \textbf{55.2} & \textbf{37.0} & \textbf{51.2} & \textbf{57.3} & 64.3 & \textbf{44.2} & \textbf{77.5} & \textbf{85.7} & \textbf{59.6} & \textbf{56.6} & \textbf{55.9} & \textbf{59.4} & \textbf{56.3} & \textbf{66.3} & \textbf{38.8} & 93.4 & \textbf{57.3} & \textbf{80.0} & \textbf{53.1} & 83.0 & \textbf{53.4} & \textbf{63.1} & \textbf{51.6} & \textbf{62.3} \\
        \bottomrule
        \end{tabular}
    }
    \label{tab:my_label}
\end{table*}

\noindent\textbf{Comparisons.} We compare against DGCNN trained under fully supervision setting (Ful. Sup.), and previous weakly supervised approaches (Weak. Sup.). Among weakly supervised approaches, we compare with WeakSup~\cite{xu2020weakly} and One Thing One Click (OTOC)~\cite{liu2021one}. For a fair comparison with OTOC, we modify the encoder of One Thing One Click with DGCNN encoder network and retain the moving average update of memory bank/prototypes. The resulting method is thus termed OTOC*. Finally, we evaluate our multi-prototype classifier (MulPro) under the same settings.

\noindent\textbf{Evaluation Metric.} We calculate the mean Intersect over Union (mIoU) for each test sample as its evaluation metric. For ShapeNet, we present the average mIoU over all samples (SampAvg) and the average mIoU over all categories (CatAvg) which we firstly calculate the average mIoU over samples in each category. For S3DIS, we present the average mIoU over all categories (CatAvg).

\subsubsection{Quantitative Results for Semantic Segmentation}

\noindent\textbf{ShapeNet Part Segmentation.} We present the results on ShapeNet part segmentation in Tab.~\ref{tab:ShapeNet}. We make the following observations from the results. i) Our model outperforms other weakly supervised models significantly. For 1 point weak supervision, our model surpasses previous state-of-the-art model by nearly 4\% and is only 5\% lower than fully supervised approach. For 10\% weak supervision, our model demonstrates comparable to OTOC*, surpasses WeakSup by 0.3\% and is closer to the full supervision approach. ii) Improvement is more substantial at lower labeling regime, suggesting the multi-prototype classifier is particularly effective when labels are sparse. 

\noindent\textbf{S3DIS Semantic Segmentation.} We present the results on S3DIS semantic segmentation in Tab.~\ref{tab:S3DIS_PartNet}. We make following observations similarly. i) Our model outperforms other models, even comparable to fully supervised approach with only 1pt (less than 0.2\%) labelled point. For 1pt weak supervision, our model outperform WeakSup\cite{xu2020weakly} by 3\% mIoU. For 10\% weak supervision, our model surpasses the previous state-of-the-art model by 0.8\% and is better than fully supervised approach. ii) Most of the categories have seen a big boost with few labeled data thanks to the multi-prototype classifier. 

\noindent\textbf{PartNet Semantic Segmentation.} We present semantic segmentation results on PartNet segmentation task. We evaluate our MulPro final model with post-processing technique~\cite{xu2020weakly}. We also compare our results with WeakSup~\cite{xu2020weakly}, the large margin suggests the efficacy of proposed mutli-prototype classifier.

\subsubsection{Qualitative Results for Semantic Segmentation}

\begin{figure*}
    \centering
    \includegraphics[width=0.98\linewidth]{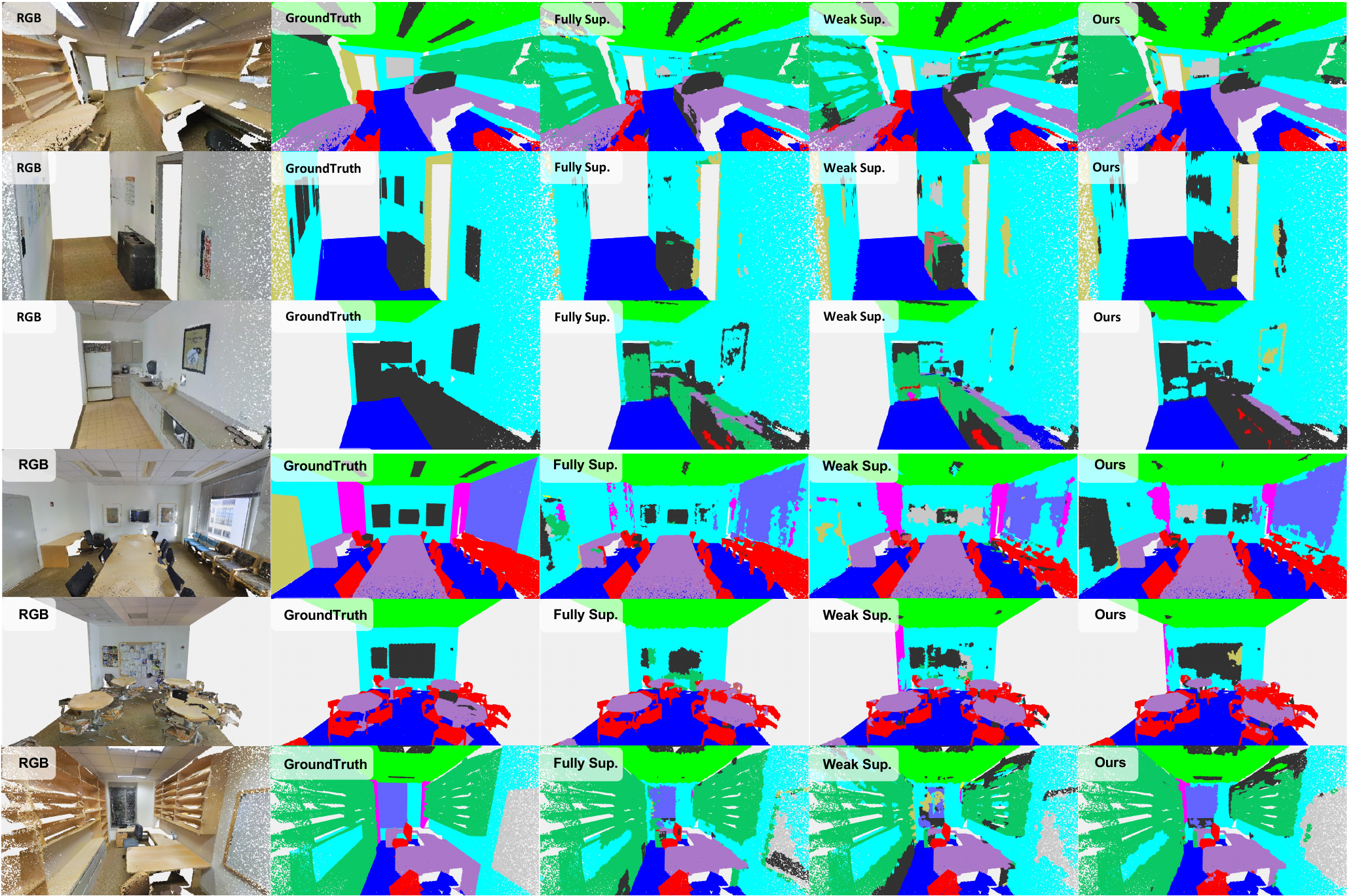}
    \caption{Qualitative examples for S3DIS semantic segmentation. Weak Sup. refers to the results of~\cite{xu2020weakly}.}
    \label{fig:S3DIS_Qualitative}
\end{figure*}

\begin{figure*}
    \centering
    \includegraphics[width=0.98\linewidth]{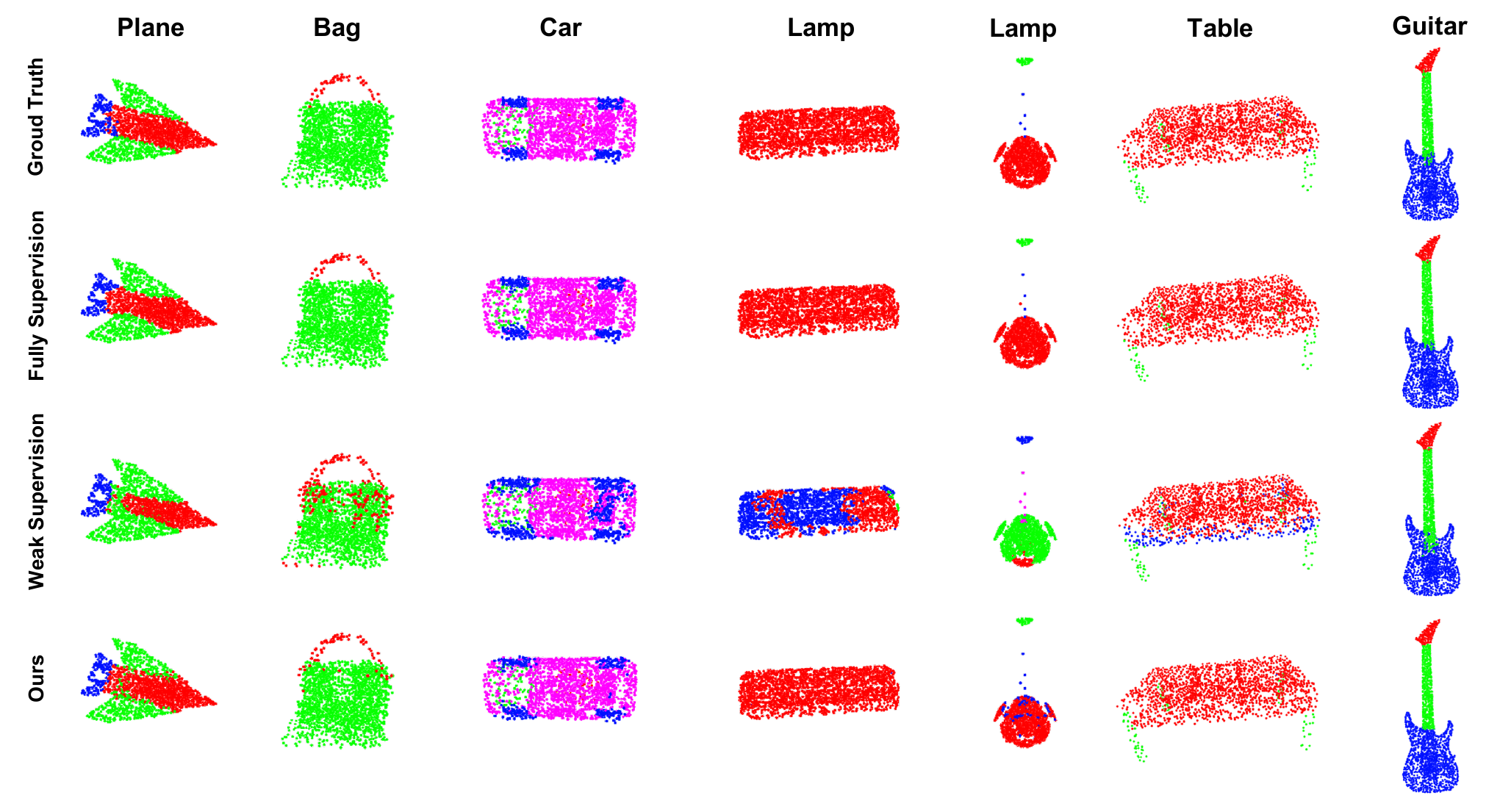}
    \caption{Qualitative examples for ShapeNet part segmentation. Weak supervision refers to the method of \cite{xu2020weakly}.}
    \label{fig:ShapeNetExample}
\end{figure*}

We show qualitative results of point cloud segmentation and compare the segmentation quality. 
For S3DIS dataset, we visualize selected segmentation samples in Fig.~\ref{fig:S3DIS_Qualitative}. From left to right, the RGB view, ground-truth, fully supervised segmentation, WeakSup\cite{xu2020weakly} segmentation and our MulPro result are visualized. In these visualization results, both MulPro and WeakSup leverages 1pt labelled points in the training stage. 
{We observe that our results better respect the ground-truth for classes with large intra-class variation. For example, a ``clutter'' category (in black color) exists in S3DIS which covers multiple types of objects that do not fall into the other 12 predefined classes. Because of the multi-prototype classifier, our model is able to identify subclasses within this ``clutter'' category. This is reflected by the more consistent predictions for ``clutter'' class. In contrast, WeakSup makes more erroneous predictions on the ``clutter'' class.}
For ShapeNet, we show the segmentation results in Fig.~\ref{fig:ShapeNetExample}. 
{These examples again demonstrate competitive performance by our model when facing categories with large intra-class variation, such as the examples of the car, lamp and plane.}

\begin{figure*}
    \centering
    \includegraphics[width=1\linewidth]{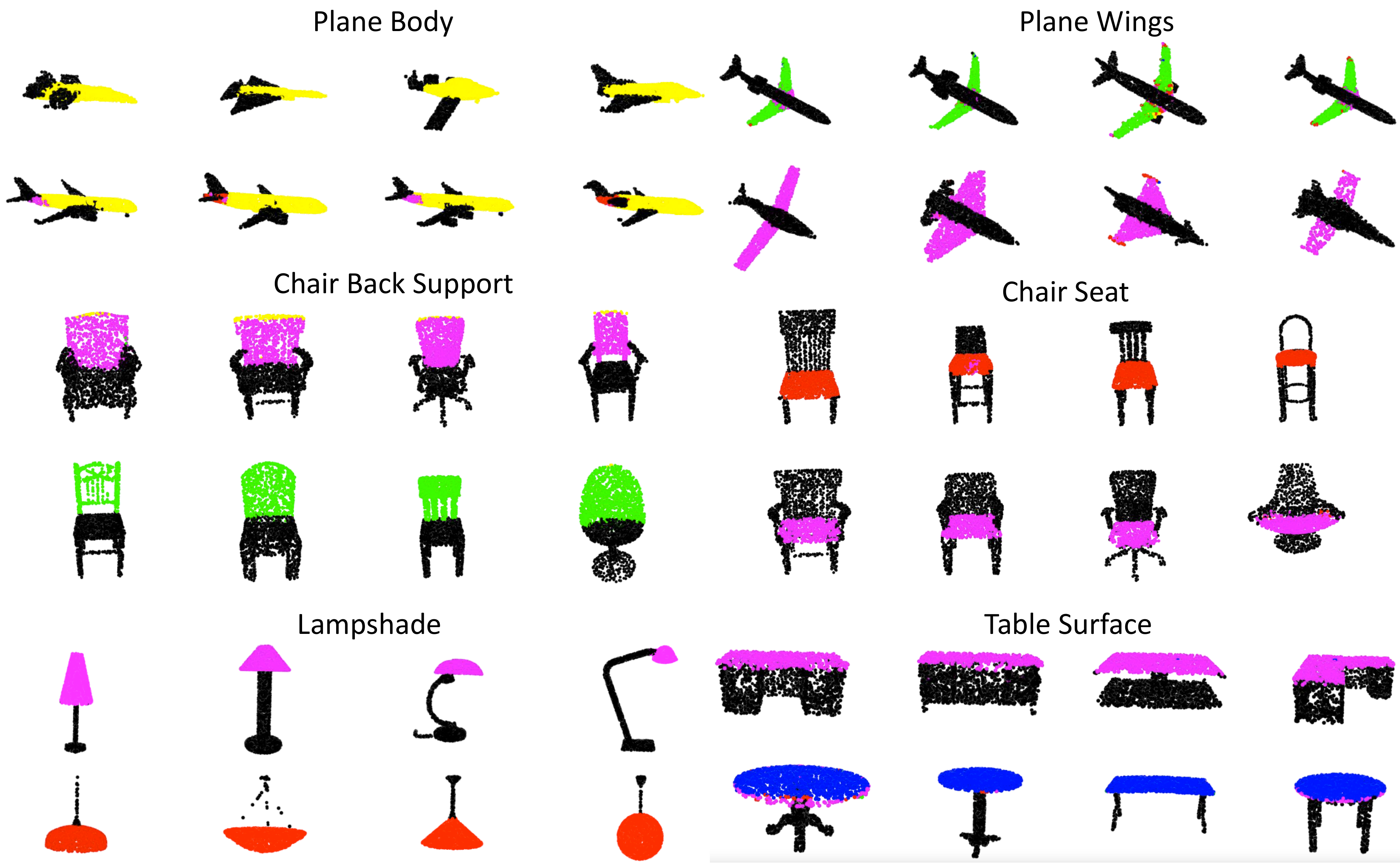}
    \caption{Visualization of activated prototypes (indicated by different colors) on selected samples from ShapeNet dataset. 1pt labelled points are used to train the weak supervision model. }
    \label{fig:Prototypes}
\end{figure*}

\subsection{Multi-Prototype Classifier Analysis}\label{subsection:multiPrototypeClassifierAnalysis}

\noindent\textbf{Discovering Subclasses.}
Multi-prototype classifier is motivated by the subclass structures within a semantic class. 
In this section, we provide qualitative results on the subclasses discovered by MulPro. In specific, for each point we define the corresponding activated prototype following Eq.~(\ref{eq:activeproto}). Given a maximal $M$ prototypes for each category we can thus assign each point into one of the prototype by its activation. In Fig.~\ref{fig:Prototypes}, we selectively visualize points by the activated prototype, i.e. each individual color indicates one activated prototype. We are surprised to see many subclasses identified. For instance, despite a single \textit{body} class is annotated for all planes, our multi-prototype classifier discovers an additional subclasses marked as red points in Fig.~\ref{fig:Prototypes}, corresponding to the tail part of \textit{body} which is shared by all passenger jets but absent from fighter jets. Two types of wings are also discovered from the plane category, roughly differentiating passenger jets from fighters. Subclasses are also discovered from chairs, different \textit{back support}s and \textit{seat}s are discovered, roughly distinguishing chairs with armrest from others. Subclasses are identified for \textit{lampshade} as well, the ones in red are generally pendants while the pink ones are mostly floor lamps. The \textit{table surface} again displays subclass structures with square-shaped desks being identified from round-shaped tables. The consistent and clean activation of prototypes among all semantic objects implies an obvious subclasses structures in the feature space.

\subsection{Multi-Prototype Classifier at Higher Label Budget}
{Multi-prototype captures the intra-class variance and is motivated under the weakly supervised setting}. As demonstrated by many research on large-scale datasets with large intra-class variation, e.g ImageNet, neural network is able to model the multi-modal distribution and a single prototype is enough for classification. However, {when labeled data points are few the neural network would face underfitting}, i.e. it fails to capture the complex data distribution and can not group features in tight clusters. Therefore, multiple prototypes are necessary at weakly supervised learning.

To validate the advantage of MulPro, we carry out additional experiments at $100\%$ labeled data. The results in Tab.~\ref{tab:mulpro_higherbudget} clearly suggests multi-prototype is most effective at 1pt annotation and the gap between DGCNN and MulPro diminishes at higher labeling regime. This observation also motivates us to explore multiple prototypes in weakly supervised setting.

\begin{table}[htbp]
    \centering
    \caption{Comparing MulPro at different labeling budgets with DGCNN as backbone.}
    \resizebox{0.9\linewidth}{!}{
        \begin{tabular}{c|c|cc}
        \toprule
        Method & Annotation & SampAvg(\%) & CatAvg(\%) \\
        \midrule
        DGCNN & 1pt & 72.6 & 72.2 \\
        DGCNN & 10\% & 84.5 & 81.5  \\
        DGCNN & 100\% & 85.1 & 82.3 \\
        \midrule
        DGCNN + MulPro & 1pt & 79.4 & 77.8 \\
        DGCNN + MulPro & 10\% & 85.3 & 82.0 \\
        DGCNN + MulPro & 100\% & 85.5 & 82.4 \\
        \bottomrule
        \end{tabular}
    }
    \label{tab:mulpro_higherbudget}
\end{table}

\subsection{Ablation and Additional Study} \label{subsection:ablationStudy}
\subsubsection{Importance of Individual Components.}
We analyze the importance of the proposed modules. Different combinations of the modules are evaluated on ShapeNet brenchmark dataset with 1pt annotation scheme. The results are presented in Tab.~\ref{tab:Ablation}. We notice that the prototype diversity must simultaneously incorporate Eq.~(\ref{eq:protodiv}) and Eq.~(\ref{eq:MIMI}) to avoid trivial solution and we combine them in the ablation study. From the ablation results, 
we first evaluate  multi-prototype classifier alone, and it yields slightly worse result than baseline, suggesting multi-prototype cannot be effectively trained with cross-entropy alone. Then we combine multi-prototype classifier with {subclass averaging} loss, this improves upon multi-prototype alone, indicating multi-prototype requires more supervision than cross-entropy loss to update. Finally, we combine prototype diversity terms and demonstrate the best result. We also evaluate our results using the post-processing technique (w/PP) proposed in \cite{xu2020weakly} which implements label propagation on network prediction and observe further improvement.



\subsubsection{Compatibility with Alternative Backbone}
We further evaluate the multi-prototype classifier with state-of-the-art 3D point cloud backbone network, namely the Point Cloud Transformer (PCT)~\cite{guo2021pct} and present the results at 1pt annotation on S3DIS dataset in Tab.~\ref{tab:PCT}. Significant improvement is observed by combining PCT with Multi-prototype classifier.

\begin{table}[htbp]
    \centering
    \caption{The results of Point Cloud Transformer encoder on S3DIS dataset.}
    \resizebox{0.65\linewidth}{!}{
        \begin{tabular}{c|c|c}
        \toprule
        Method & Annotation & mIoU (\%) \\
        \midrule
        PCT~\cite{guo2021pct} & 1pt & 41.6 \\
        PCT + MulPro & 1pt & 43.0 \\
        \midrule
        PCT & 100\% & 51.5 \\
        \bottomrule
        \end{tabular}
    }
    \label{tab:PCT}
\end{table}

\subsubsection{Number of Multi-Prototype}
\begin{figure}
    \centering
    \subfloat[Number of available prototypes]{\includegraphics[width=0.5\linewidth]{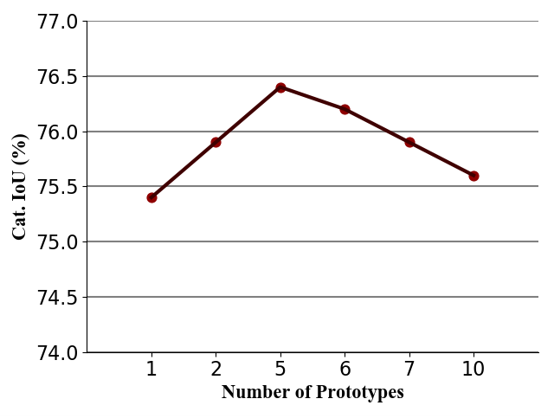}}
    \subfloat[Number of activated prototypes]{\includegraphics[width=0.5\linewidth]{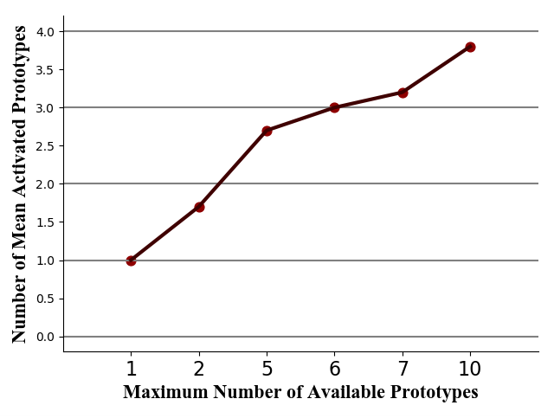}}
    \caption{The impact of \#prototypes on ShapeNet dataset. (a) shows the change of mIoU with the number of prototypes; (b) shows the change of the mean number of activated prototypes with the number of prototypes.}
    \label{fig:NumMultiProto}
\end{figure}

We investigate the impact of the number of multi-prototype on segmentation performance. In specific, we evaluate $M=1\cdots 10$ on ShapeNet segmentation with 1 point per category label. We present the results in Fig.~\ref{fig:NumMultiProto} from which we observe that 
the number of activated prototypes increases with the increase of the number of available prototypes ($M$), however, the mean category IoU reaches the maximum value when $M = 5$.

\begin{table}[!t]
    \centering
    \caption{Ablation study on the impact of individual modules. The results consist of without Post Processing (wo/PP) and with Post Processing (w/PP).}
        \setlength\tabcolsep{1pt} 
    \resizebox{0.99\linewidth}{!}{
        \begin{tabular}{ccc|cc}
        \toprule
        \multicolumn{3}{c|}{Components} & \multicolumn{2}{c}{mIoU (\%)} \\
        \midrule
        Multi-Prototype & Subclass Averaging &  Prototype Diversity & w/o PP & w/ PP~\cite{xu2020weakly} \\
        \cmidrule(lr){1-1} \cmidrule(lr){2-2} \cmidrule(lr){3-3} \cmidrule(lr){4-4} \cmidrule(lr){5-5}
        - & - & - & 73.8 & 74.4 \\
        \checkmark & - & - & 73.7 & - \\
        \checkmark & \checkmark & - & 75.4 & - \\   
        \checkmark & \checkmark & \checkmark & \textbf{76.4} & \textbf{77.8} \\
        \bottomrule
        \end{tabular}
    }
    \label{tab:Ablation}
\end{table}

\subsubsection{Subclass Averaging Constraint Design.}\label{subsection:ReconLossDesign}
Due to the importance of subclass averaging loss, we investigate several alternative designs. 
First, an naive way to use both labeled and unlabeled data to update prototype is through pseudo-labeling~\cite{iscen2019label}. We predict the pseudo-labels for all unlabeled data points and the pseudo-labels are in turn used to supervised cross-entropy loss on unlabeled data.
Alternatively, we could use Frobenius norm to measure the distance between data points and corresponding activated prototypes. Specifically, we stack all activated prototypes over data points as, $\hat{\Omega}=[\vect{\omega}_{\hat{m}_1\hat{k}_1}; \cdots ;\vect{\omega}_{\hat{m}_N\hat{k}_N}] \in \mathbb{R}^{D_o\times N}$. Then the subclass averaging loss is calculated as, $\mathcal{L}_{avg1} = ||\hat{\Omega}-\matr{Z}||_F^2$. 
Since the distance metric consists of Frobenius norm and inner product between $\hat{\Omega}$ and $\matr{Z}$, the Frobenius norm parts will affect the scale, it might lead to trivial solutions. To avoid these trivial solutions, we propose the second candidate, using cosine distance to perform subclass averaging loss as,
\begin{equation}
    \mathcal{L}_{avg2} =-\sum_n{\frac{\langle \vect{\omega}_{\hat{m}_n \hat{k}_n}, \vect{z}_n \rangle}{||\vect{\omega}_{\hat{m}_n \hat{k}_n}|| \cdot ||\vect{z}_n||}}
\end{equation}

Results for comparing all alternative designs are presented in Tab.~\ref{tab:ReconstructionLoss}. 
We make the following observations from the results. Firstly, pseudo-labeling gives the worst results, probably due to the confirmation bias~\cite{arazo2020pseudo} to blame.
Furthermore, using the cosine similarity to update prototypes $\mathcal{L}_{avg2}$ avoids the trivial solution and the performance is slightly better than that using L2 distance alone $\mathcal{L}_{avg1}$. Finally, our thresholded subclass averaging outperforms both candidates, suggesting it is necessary to selectively use most relevant point features to update prototypes.

\begin{table}[htbp]
    \centering
    \caption{The results of different options of the {subclass averaging} loss.}
    \resizebox{0.9\linewidth}{!}{
        \begin{tabular}{c|cc}
        \toprule
        Subclass Averaging Options & ShapeNet & S3DIS \\
        \midrule
         Pseudo-Labeling & 74.2 & 44.4 \\
        $\mathcal{L}_{CE}+\mathcal{L}_{avg1}+\mathcal{L}_{pd} + \mathcal{L}_{bds}$ & 74.8 & 44.9 \\
        $\mathcal{L}_{CE}+\mathcal{L}_{avg2} + \mathcal{L}_{pd} + \mathcal{L}_{bds}$ & 75.8 & 45.5 \\
        $\mathcal{L}_{CE}+\mathcal{L}_{avg} + \mathcal{L}_{pd} + \mathcal{L}_{bds}$ (Ours) & \textbf{76.4} & \textbf{46.8} \\
        \bottomrule
        \end{tabular}
    }
    \label{tab:ReconstructionLoss}
\end{table}

\begin{table*}[!htb]
    \centering
    \caption{The numbers of activated prototypes in each shape on ShapeNet dataset.}
    \resizebox{0.99\linewidth}{!}{
        \begin{tabular}{c|cccccccccccccccc}
        \toprule
        Subclass Averaging Options & Air. & Bag & Cap & Car & Chair & Ear. & Guitar & Knife & Lamp & Lap. & Motor. & Mug & Pistol & Rocket & Skate. & Table \\
        \midrule
        $\mathcal{L}_{CE}+\mathcal{L}_{avg1}+\mathcal{L}_{pd} + \mathcal{L}_{bds}$ & 1 & 1 & 1 & 1 & 1 & 1 & 1 & 1 & 1 & 1 & 1 & 1 & 1 & 1 & 1 & 1 \\
        $\mathcal{L}_{CE}+\mathcal{L}_{avg2} + \mathcal{L}_{pd} + \mathcal{L}_{bds}$ & 1 & 1 & 1 & 1 & 1 & 1 & 1 & 1 & 1 & 1 & 1 & 1 & 1 & 1 & 1 & 1 \\
        $\mathcal{L}_{CE}+\mathcal{L}_{avg} + \mathcal{L}_{pd} + \mathcal{L}_{bds}$ (Ours) & \textbf{5} & \textbf{1} & \textbf{1} & \textbf{5} & \textbf{5} & \textbf{1} & \textbf{3} & \textbf{3} & \textbf{5} & \textbf{1} & \textbf{1} & \textbf{3} & \textbf{2} & \textbf{1} & \textbf{1} & \textbf{5} \\
        \bottomrule
        \end{tabular}
    }
    \label{tab:ActivatePrototypeShapeNet}
\end{table*}

\begin{table*}[htb]
    \centering
    \caption{The numbers of activated prototypes in each semantic part on S3DIS (Area 5) dataset.}
    \resizebox{0.99\linewidth}{!}{
        \begin{tabular}{c|ccccccccccccc}
        \toprule
        Subclass Averaging Options & ceil. & floor & wall & beam & col. & win. & door & chair & table & book. & sofa & board & clutter \\
        \midrule
        $\mathcal{L}_{CE}+\mathcal{L}_{avg1}+\mathcal{L}_{pd} + \mathcal{L}_{bds}$ & 1 & 1 & 5 & 1 & 1 & 1 & 2 & 1 & 2 & 1 & 1 & 1 & 6\\
        $\mathcal{L}_{CE}+\mathcal{L}_{avg2} + \mathcal{L}_{pd} + \mathcal{L}_{bds}$ & 1 & 1 & 5 & 1 & 1 & 1 & 3 & 3 & 2 & 1 & \textbf{10} & 1 & 5\\
        $\mathcal{L}_{CE}+\mathcal{L}_{avg} + \mathcal{L}_{pd} + \mathcal{L}_{bds}$ (Ours) & \textbf{9} & \textbf{7} & \textbf{7} & \textbf{2} & \textbf{4} & \textbf{1} & \textbf{6} & \textbf{3} & \textbf{6} & \textbf{2} & 4 & \textbf{2} & \textbf{6} \\ 
        \bottomrule
        \end{tabular}
    }
    \label{tab:ActivatePrototypeS3DIS}
\end{table*}





We further compare the number of uniquely activated prototypes under different subclass averaging losses. We present the results on ShapeNet and S3DIS in Tab.~\ref{tab:ActivatePrototypeShapeNet} and Tab.~\ref{tab:ActivatePrototypeS3DIS} respectively.
We make the following observations from the results. First, the numbers of activated prototypes are very few with both L2 distance and cosine distance as subclass averaging loss on both ShapeNet and S3DIS. This is probably due to the challenge in prototype initialization. When one or a few prototypes are activated there is no force to encourage the activation of other ones.
Moreover, with our proposed subclass averaging loss, we notice certain classes have more activated prototypes, e.g. ``airplane'', ``car'', ``chair'' on ShapeNet and ``ceil.'', ``door'', ``chair'', ``table'', ``clutter'' on S3DIS, suggesting large intra-class variation. 

\section{Conclusion}
In this work, we first observe that existing approaches towards point cloud segmentation often employ a linear classifier to separate semantic classes. This is equivalent to allocating one prototype for each category. As we point out through experiment and intuition, clear subclass structures in each semantic class of 3D point cloud segmentation data widely exist and would result in large intra-class variation in feature representation. As a result, the single prototype may not capture the large variation and lead to inferior results. To tackle this issue, we proposed a multi-prototype memory bank where each prototype serves as the classifier for one subclass. To enable effective multi-prototype training, we further introduced two constraints. Extensive results on weakly supervised 3D point cloud segmentation benchmark suggest the advantage of maintaining multiple prototypes in particular at low-label regime. We hope the subclasses identified from ShapeNet could provide insights into future segmentation model design at low-label regime.



%
%
%
%
%
%
%
%

\ifCLASSOPTIONcaptionsoff
  \newpage
\fi



%
\bibliographystyle{IEEEtran}

\bibliography{reference}
%
%
%
%
%




\end{document}